\documentclass[runningheads]{llncs}

\usepackage[T1]{fontenc}
\usepackage{graphicx}
\usepackage{color}
\usepackage{url}
\usepackage{amsmath}
\usepackage{booktabs}

\begin{document}
\title{Patient-Level Elbow Abnormality Detection: Leakage-Aware Evaluation of Learned Preprocessing, Calibration, and Triage-Oriented Operating Points}
\titlerunning{Patient-Level Elbow Abnormality Detection}

\author{Ahmed Sallam\inst{1}\orcidID{0000-0002-4864-1275} \and
Ahmet Kaplan\inst{2}\orcidID{0000-0001-5231-2282}}

\authorrunning{A. Sallam and A. Kaplan}

\institute{Biomedical Engineering Department, Istanbul Medipol University, Istanbul, Türkiye\\
\email{ahmed.sallam@std.medipol.edu.tr} \and
Computer Engineering Department, Istanbul Medipol University, Istanbul, Türkiye\\
\email{ahmet.kaplan@medipol.edu.tr}}

\maketitle

\begin{abstract}

In this study, we examine learned preprocessing pipelines in the context of triage-oriented orthopedic abnormality detection task using elbow radiographs from MURA dataset. The evaluation focuses on patient-level detection of musculoskeletal abnormalities under a leakage-aware protocol. We compare multiple preprocessing pipelines, with and without a lightweight DnCNN module as a learned preprocessing component, to assess their impact on discrimination and calibration. Performance is assessed using discrimination metrics (AUROC, PR-AUC), calibration measures (ECE, Brier score), and validation-selected operating point analysis targeting high specificity. Results show that differences across preprocessing strategies are modest and configuration-dependent, with no consistent discrimination advantage over the raw-input DenseNet121 baseline. The raw and diverse inputs combined with the DnCNN front-end showed reduced ECE and Brier score, while CLAHE combined with DnCNN did not improve calibration. Overall, the results suggest that under patient-level evaluation, preprocessing gains are modest and configuration-dependent; the raw-input DenseNet121 baseline remains competitive throughout, and no tested preprocessing strategy produced a consistent discrimination advantage across all metrics.

\end{abstract}

\keywords{Musculoskeletal abnormality detection  \and orthopedic triage \and Learned preprocessing  \and DnCNN  \and Patient-level evaluation  \and Elbow radiographs  \and Triage-oriented AI}

\section{Introduction}

One of the leading causes of emergency department visits is bone fractures and musculoskeletal abnormalities, which place a burden on clinical decision-making in the emergency department \cite{Pinto2018}. In addition to accurate diagnosis, it is necessary to triage cases rapidly. Evidence suggests that specialized orthopedic and musculoskeletal triage protocols enhance clinical efficacy and reduce ER length of stay\cite{ROBINSON2013398,samsson2020effects}.

Convolutional neural networks (CNNs) have shown promising results in musculoskeletal abnormality detection from radiographs\cite{rajpurkar2018}. Multiple studies have explored the effectiveness of DenseNet-based architectures in medical imaging and have become a robust baseline in the domain \cite{app10041507,Tahir2024,Meena2022}. Nevertheless, meta-analysis and systematic reviews pointed out many limitations on the clinical application of such models\cite{Husarek2024,RAOKARANAM20232557}. These limitations include the lack of strict patient-level dataset splits, which can lead to data leakage between training and evaluation.

Although much recent evidence has shown that AI-based fracture detection systems are achieving high accuracy, many have reported a lack of generalizability and reliability. A systematic review and meta-analysis reviewed 42 studies; on internal validation, it reported a pooled sensitivity and specificity of 92\% and 91\%, respectively, and 91\% on external validation, with no improvement over clinical performance \cite{kuo_artificial_2022}. On an important note, just 13 studies tested external validation, and 22 studies are judged to have high risk of bias, which suggests lack of generalizability to real-life clinical settings. To further emphasize this point, another systematic review and meta-analysis analyzed 66 studies and reported bias via patient selection and reference standards, which raises concerns of real-life applicability \cite{jung_artificial_2024}.

The literature suggests that fracture-AI performance is subject to variation in anatomy and use cases. A systematic review observed a robust diagnostic accuracy across the majority of the evaluated AI tools, peaking when AI is combined with human; however, a lower accuracy was observed for spinal and rib fractures \cite{Husarek2024}. Additionally, a multi-site study demonstrated the capability of deep-learning in accurately detecting the presence of fracture in the adult musculoskeletal system \cite{jones_assessment_2020}.

Taken together, the potential of AI for fracture detection is established, but when considering a deployment-oriented use case, rigorous and careful evaluation is essential.

The enhancement of medical images plays a critical role in improving their visibility. Image reprocessing provides various techniques that help improve contrast, brightness, and global cues, as well as removing and suppressing noise and irrelevant information. Contrast Limited Adaptive Histogram Equalization (CLAHE) is one of the widely used techniques in medical imaging \cite{Luo_2021}. In recent years, learned preprocessing techniques have become popular as they have the capability to suppress irrelevant noise and sustain meaningful cues, which can help the network in learning. A recent study suggested that targeted preprocessing strategies have the potential to improve classification performance by shifting the model's attention to specific regions that contain abnormality cues \cite{wu_enhancing_2025}. This supports the idea that preprocessing is a major variable in the whole process that is worth experimenting with and controlled evaluation.

Zhang et al introduced a denoising convolutional neural network (DnCNN), which is a residual-learning-based convolutional neural network designed for image denoising \cite{Zhang_2017}. This technique has shown good performance by learning to predict the noise rather than cleaning the signal directly. The following work has examined the usage of DnCNN in the context of medical imaging, including the enhancement of radiological images \cite{9137813,Kangralkar2025}. Regardless of DnCNN's usefulness as a standalone denoiser, the integration of DnCNN as a preprocessing pipeline with a DenseNet backbone is still not fully explored.

In this study, we focus on orthopedic triage-oriented evaluation by examining musculoskeletal abnormality detection at the patient level, considering both discrimination and the reliability of predicted probabilities. By employing the elbow subset of the MURA dataset and comparing multiple preprocessing pipelines, 

with and without a lightweight DnCNN-based learned preprocessing module, while keeping a DenseNet121 backbone fixed across experiments. We analyze calibration behavior using expected calibration error (ECE) and Brier score, and evaluate clinically motivated operating points targeting high specificity. This allows us to examine how predicted probabilities behave under triage-oriented operating conditions.

The main contribution of this study is in four parts. First, we implement a patient-level, leakage-aware evaluation protocol for elbow abnormality detection on the MURA dataset with a deterministic dataset splitting. Second, we compare classical and learned preprocessing pipelines under controlled backbone settings using DenseNet121 across all experiments. Third, we report discrimination metrics and calibration metrics and analyze probability reliability at the patient level. Fourth, we examine validation-selected high-specificity operating points on the test set to reflect a conservative triage-oriented evaluation scenario. Specifically, this study performs a controlled evaluation of preprocessing pipelines under patient-level dataset splitting, multi-seed experimental repetition, calibration analysis, and triage-oriented operating-point evaluation.

\section{Methods}

\subsection{Dataset and curation}

In this study, we used the elbow subset of the publicly available MURA dataset \cite{rajpurkar2018}. In order to exclude images with severe artifacts, we performed a manual quality control. The exclusion is based on the presence of substantial motion blur, contrast issues, incomplete visualization of the elbow joint, severe cropping of anatomical structures, or other acquisition artifacts. After the QC, we retained a total of 3,860 elbow radiographs. The dataset was then split into training (70\%), validation (15\%), and test (15\%) using a patient-level split with a deterministic random seed (42) to prevent data leakage between subsets.

The final dataset configuration was:

\begin{itemize}

    \item Training set: 2,679 images (941 patients)

    \item Validation set: 569 images (202 patients)

    \item Test set: 568 images (201 patients)

\end{itemize}

Through stratification, we maintained an approximately balanced distribution (1:1) between both classes (normal and abnormal).

\subsection{Preprocessing Pipelines}

In this study, three preprocessing pipelines were evaluated to examine whether preprocessing choice materially affects patient-level discrimination, calibration, and triage-oriented operating behavior.

\begin{itemize}

\item \textbf{Raw input:} this is limited to intensity normalization to the range of $[0,1]$. Then the grayscale radiograph image is replicated across the input channels, which is the expected input format of DenseNet networks. This is basically our baseline model.

\item \textbf{CLAHE:} We implemented Contrast Limited Adaptive Histogram Equalization (CLAHE) on the input image to enhance the local contrast. Preprocessed images were replicated across the three channels just like in raw inputs. The rationale behind using CLAHE is to highlight bone edges to enhance global cues.

\item \textbf{Diverse representation:} In this pipeline, we combined three methods: raw in one channel, CLAHE in another, and a detail-enhanced channel in the third. The detail channel is derived from a bilaterally filtered base–residual with sigmoid-gated gradient amplification: the input $I$ is decomposed into a bilateral-filter smoothed base $B = \mathrm{BF}(I;\,d{=}7,\,\sigma_c{=}20,\,\sigma_s{=}7)$ and a residual $d = I - B$. An adaptive gain $\beta = \beta_0\,\sigma\!\bigl(k(\hat{g} - 0.5)\bigr)$, where $\hat{g}$ is the normalized gradient magnitude $\|\nabla I\|$, $\beta_0{=}1$, and $k{=}4$, amplifies the residual in high-gradient regions to produce the output $d\,(1+\beta)$, normalized to $[0,1]$. The hypothesis is that exposing the model to these complementary representations simultaneously helps it learn features from different aspects of the radiograph.

\end{itemize}

We implemented a set of data augmentations to help improve the model's generalization. This includes horizontal flips, small rotations, and affine transformations with limited translation and shear. Each preprocessing configuration is evaluated across multiple random seeds.

\subsection{Model architecture}

We employed the DenseNet121 as the backbone in all our experiments. The model is a pretrained model on the ImageNet dataset. The rationale of this selection is its robust performance on medical imaging tasks and its simplicity. To investigate the impact of learned denoising, we incorporated a lightweight DnCNN module as a \emph{learnable preprocessing front-end} placed before the DenseNet121 backbone and trained end-to-end with it \cite{Zhang_2017}. It operates directly on the input image before any backbone layers, not as a post-hoc calibration or correction step applied after the backbone. The denoising module uses a residual learning formulation where the network predicts the noise component $n(x)$ of the input image $x$, which is then subtracted to obtain the denoised representation $\hat{x} = x - n(x)$ that is passed to the backbone.

\begin{figure}

\centering

\includegraphics[width=1\textwidth]{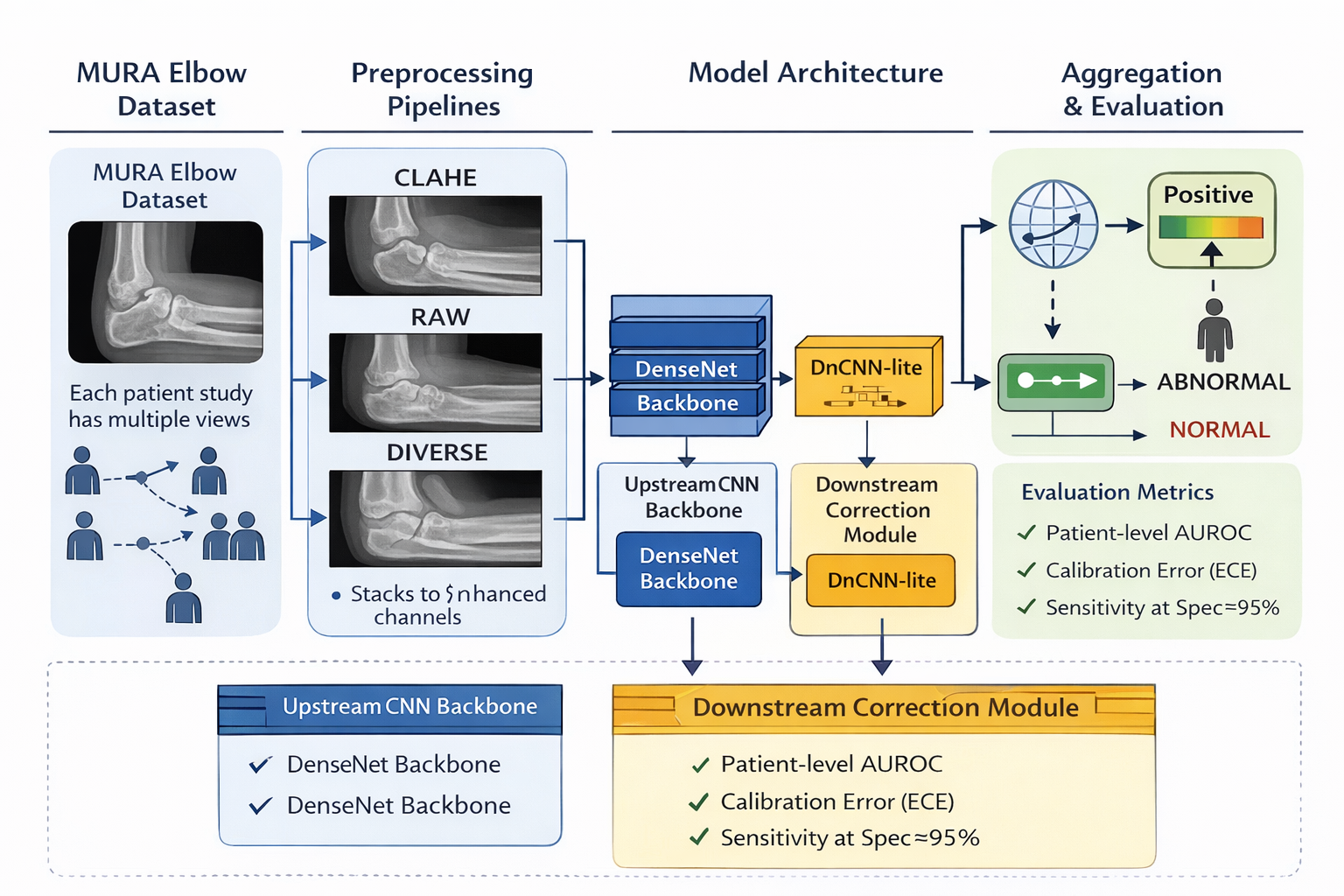}

\caption{{Overview of the proposed orthopedic triage pipeline. Patient-level studies from the MURA elbow dataset are processed using different preprocessing pipelines (raw, CLAHE, and diverse representations). A DenseNet121 backbone is employed, optionally preceded by a lightweight DnCNN module for learned denoising. Image-level predictions are aggregated at the patient level, and performance is evaluated using discrimination, calibration, and triage-oriented operating-point metrics.}}

\label{diagram}

\end{figure}

The purpose of this extension is adaptive suppression of image noise while preserving diagnostic cues. The DnCNN consists of an initial convolution layer followed by convolution-batch normalization-ReLU blocks and a final reconstruction layer. The configuration used 5 convolutional layers with 64 feature channels. We experimented with and without the DnCNN front-end to assess its contribution to each preprocessing pipeline. Figure~\ref{diagram} demonstrates the pipeline diagram, from patient-level input, preprocessing, model backbone, and evaluation.

\subsection{Training and evaluation}

Each configuration was trained using 5 random seeds to assess training stability ($n=5$, seeds $\in\{0, 42, 132, 777, 1234\}$), and results are reported as $\text{mean} \pm \text{standard deviation}$. Each seed was trained for 20 epochs. Training was performed using the MONAI framework, based on PyTorch \cite{cardoso2022}. Models were trained using AdamW optimizer with an initial learning rate of $1.1 \times 10^{-4}$ and weight decay of $1 \times 10^{-4}$.

We used binary cross-entropy with logits loss to account for label imbalance.

The high-specificity operating regime is motivated by the intended deployment role: an automated screening system that minimizes unnecessary escalation of normal cases to specialist review, while keeping sensitivity as a clinician-adjustable parameter tuned to local case-load and risk tolerance \cite{ROBINSON2013398}. Patient-level predictions are obtained by averaging image-level predicted probabilities across all views belonging to a patient. Mean aggregation is robust to variation in view count per patient and does not require a minimum number of views, making it more suitable than majority voting under our asymmetric class distribution.

Evaluation was conducted at the patient level. We used threshold-free metrics (AUROC and PR-AUC) to evaluate discrimination performance. We used ECE and Brier score on the patient level to evaluate calibration. In order to achieve target specificity levels, we conducted an operating-point analysis to simulate a triage setting. We report sensitivity, specificity, positive predictive value (PPV), and negative predictive value (NPV). The results presented are means and standard deviations across all seeds. We implemented this evaluation protocol to conduct a conservative, clinically realistic assessment of the proposed method.

\section{Results}

\subsection{Discrimination performance}

In this project, we compared preprocessing pipelines and model configurations to perform binary classification of normal and abnormal elbow studies from X-ray images at the patient level. The comparison is based on the test set threshold-free metrics, AUROC aggregated at the patient level, and then reporting the mean and standard deviation on multiple seeds. Our raw input is basically a strong baseline, which shows how the backbone model (DenseNet121) can extract meaningful features with no preprocessing besides basic intensity normalization (0 to 1). CLAHE, on the other hand, shows a slightly better discrimination in some configurations but not consistent across all settings.

We observe almost identical AUROC values across raw, CLAHE, and diverse pipelines, with slight differences at the patient level. Implementing the DnCNN module slightly improves discrimination performance on raw, diverse inputs, whereas its effect on CLAHE inputs is variable and not consistent across metrics. These results suggest that preprocessing does not outperform a raw-input DenseNet121 under patient-level, leakage-aware evaluation. Pairwise Wilcoxon signed-rank tests on patient-level AUROC across seeds confirmed that no configuration pair produced a statistically significant difference (all $p > 0.31$ before multiple-comparison correction), consistent with the narrow observed AUROC spread of 0.866--0.877. Based on this observation, we examine in greater detail the configurations that achieve the strongest performance in the following calibration and operation subsections. The small numerical differences across configurations suggest that preprocessing had a limited impact compared to the baseline backbone under the present evaluation protocol. Table~\ref{tab:supp_threshold_free} shows the AUROC and PR-AUC across all configurations.

\begin{figure}

\centering

\includegraphics[width=1\textwidth]{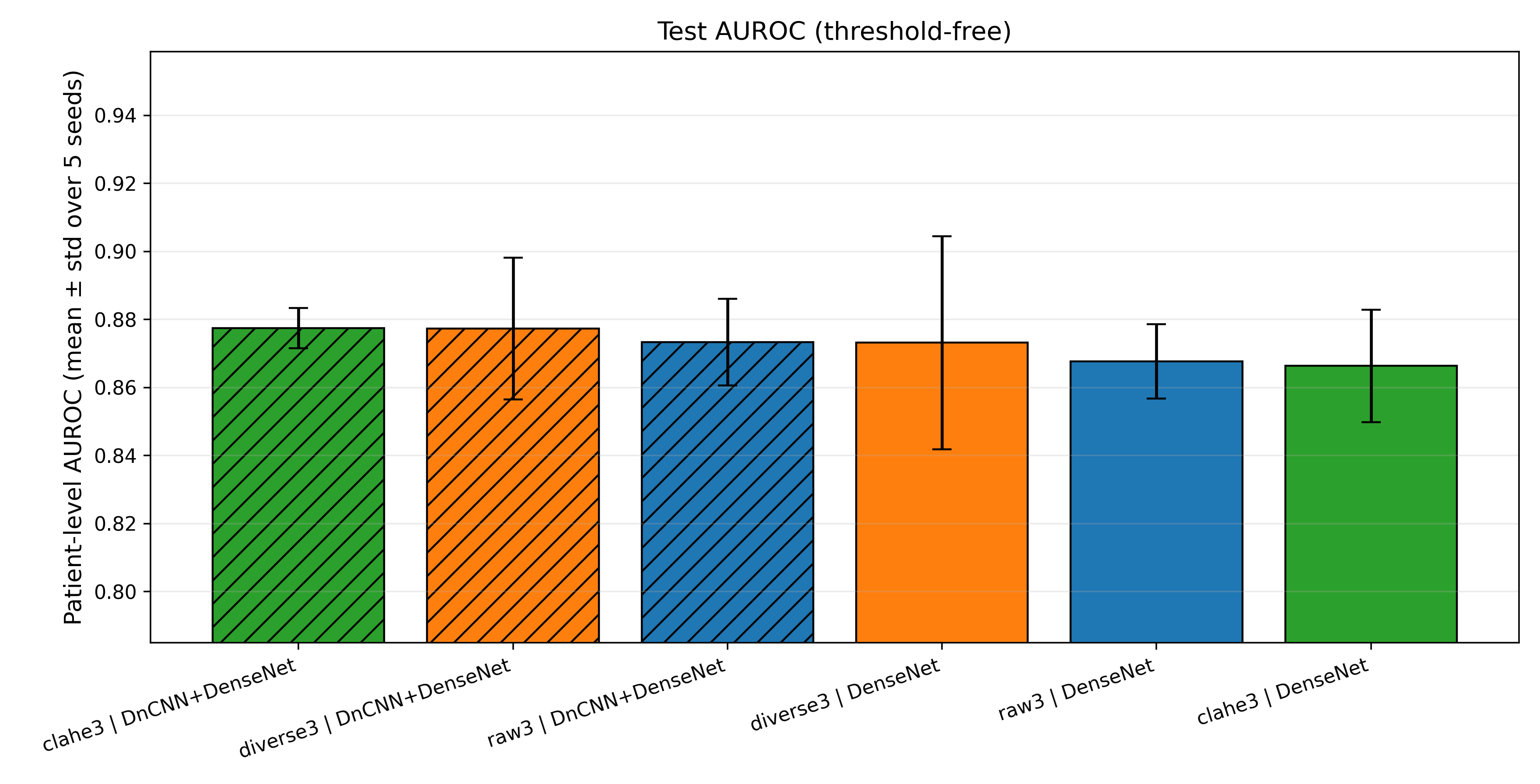}

\caption{{Patient-level AUROC on the test set. Error bars indicate standard deviation across seeds.}}

\label{fig2}

\end{figure}

\begin{table}[t]

\centering

\caption{Patient-level test-set performance (threshold-free). Values are mean $\pm$ std over seeds ($n=5$). Lower is better for ECE and Brier. Best value per column is bolded.}

\label{tab:supp_threshold_free}

\begin{tabular}{llccccc}

\toprule

Preprocessing & Variant & $n$ & AUROC $\uparrow$ & PR-AUC $\uparrow$ & ECE $\downarrow$ & Brier $\downarrow$ \\

\midrule

raw3\_nomask     & base  & 5 & 0.868 $\pm$ 0.011 & 0.891 $\pm$ 0.011 & 0.103 $\pm$ 0.016 & 0.147 $\pm$ 0.007 \\

raw3\_nomask     & dncnn & 5 & 0.873 $\pm$ 0.013 & 0.895 $\pm$ 0.015 & \textbf{0.090 $\pm$ 0.017} & \textbf{0.137 $\pm$ 0.012} \\

clahe3\_nomask   & base  & 5 & 0.866 $\pm$ 0.017 & 0.889 $\pm$ 0.023 & 0.104 $\pm$ 0.007 & 0.143 $\pm$ 0.015 \\

clahe3\_nomask   & dncnn & 5 & \textbf{0.877 $\pm$ 0.006} & 0.896 $\pm$ 0.012 & 0.116 $\pm$ 0.032 & 0.152 $\pm$ 0.023 \\

diverse3\_nomask & base  & 5 & 0.873 $\pm$ 0.031 & 0.894 $\pm$ 0.024 & 0.099 $\pm$ 0.021 & 0.152 $\pm$ 0.026 \\

diverse3\_nomask & dncnn & 5 & 0.877 $\pm$ 0.021 & \textbf{0.898 $\pm$ 0.023} & 0.100 $\pm$ 0.009 & 0.141 $\pm$ 0.022 \\

\bottomrule

\end{tabular}

\end{table}

\subsection{Calibration and probability reliability}

In a triage setting, decisions are made on the model's confidence, which is based on reliable probability estimates. Here, we evaluated the predicted probabilities at the patient level. Expected calibration error (ECE) and Brier score were used to measure calibration on the test set at the patient-level after aggregation. 

Some preprocessing pipelines that utilized DnCNN showed a lower calibration error, suggesting improved alignment between empirical outcomes and predicted probabilities in selected configurations rather than a uniform effect.

\begin{figure}

\centering

\includegraphics[width=1\textwidth]{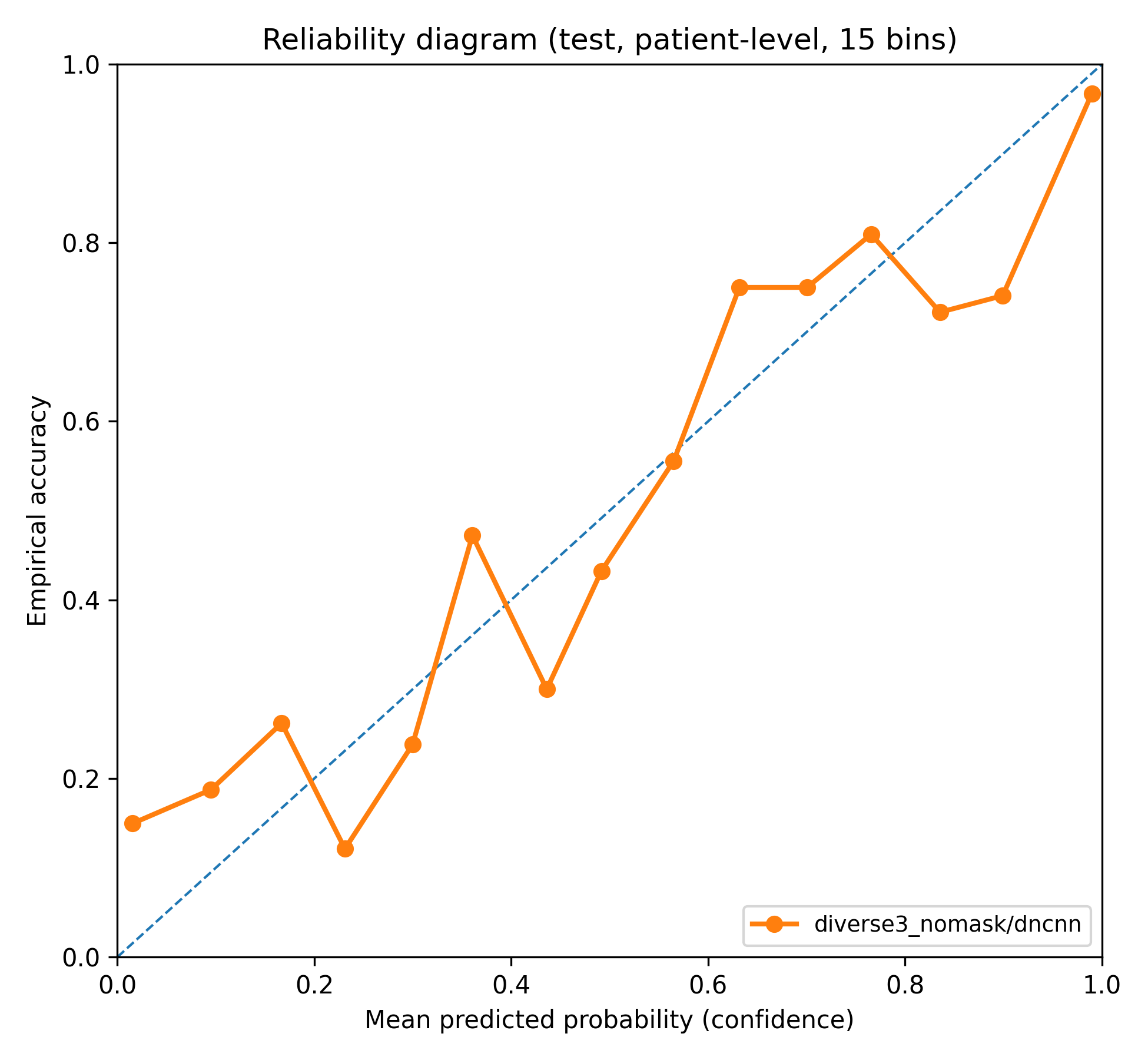}

\caption{Reliability diagram for the diverse + DnCNN configuration on the test set at the patient level ($n=201$ patients). The dashed line indicates perfect calibration. Each point represents the mean predicted probability and empirical accuracy within a confidence bin; bin populations are smallest in the mid-probability range ($0.1$--$0.5$), which accounts for the larger deviations observed there.}

\label{fig:reliability}

\end{figure}

In the proposed model, the reliability curve follows the identity line closely across the probability range on the test set at the patient level, suggesting that predicted confidence is generally consistent with observed accuracy (Figure~\ref{fig:reliability}). We can see some minor deviations in the mid-region; however, that is possibly due to the limited sample counts per bin. Mainly, we can see limited overconfidence, which is critical for triage decisions. When the model assigns a great abnormality probability, it aligns closely with the actual accuracy.

These findings suggest that shifting the focus to confidence-based thresholds can better help in a triage-oriented setting.  

This behaviour supports the idea that the approach helps create reliable estimates rather than unreliable predictions. The reliability analysis demonstrates that specific configurations produce well-calibrated patient-level probability scores, rather than the typical pitfall of severe overconfidence.

\subsection{Triage operating-points}

\begin{figure}

\centering

\includegraphics[width=1\textwidth]{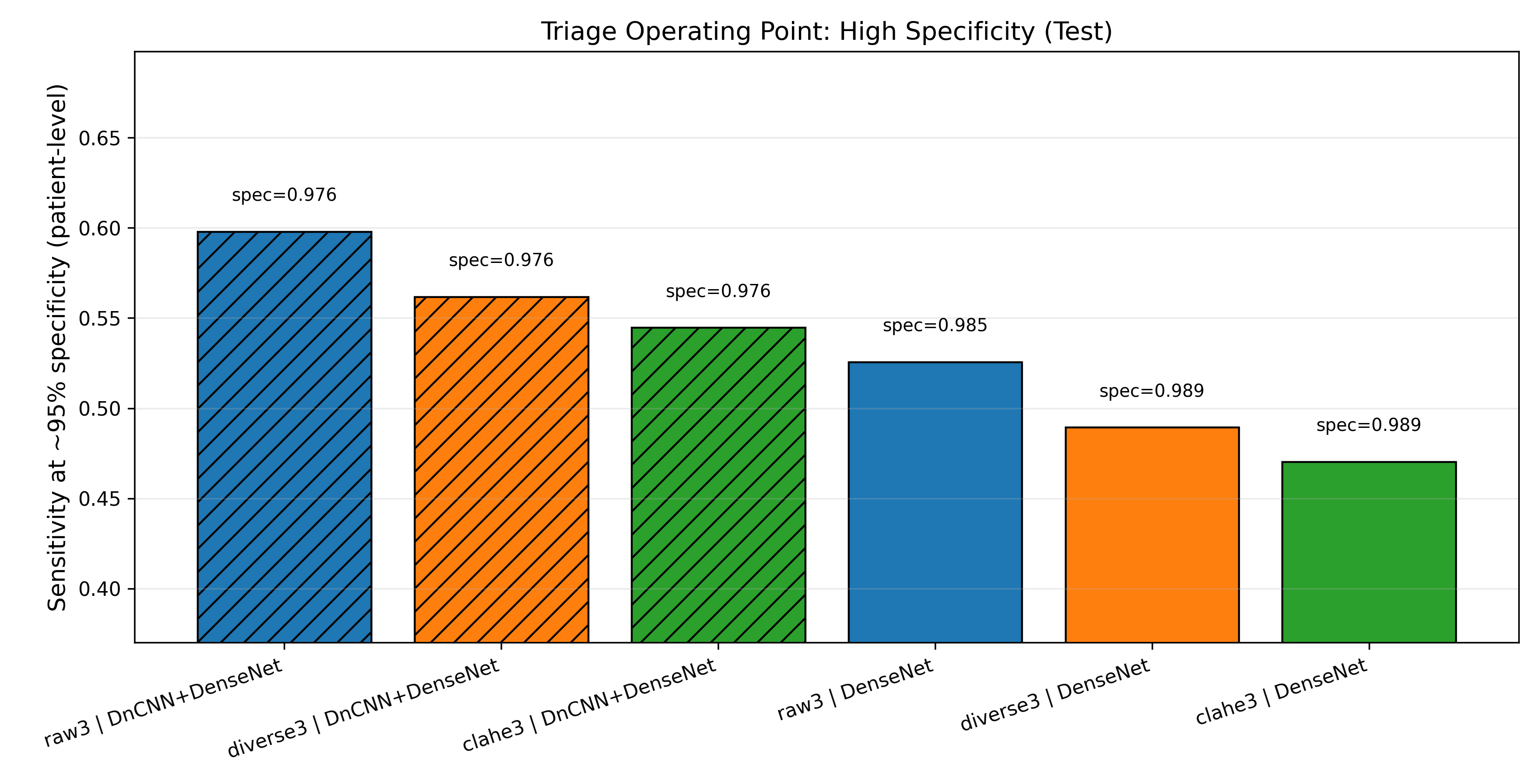}

\caption{Patient-level sensitivity at a target specificity of 0.95 for each of the six configurations on the test set. Error bars indicate standard deviation across seeds ($n=5$). Configurations with the DnCNN front-end achieve higher sensitivity at this operating point on raw and diverse inputs, while CLAHE+DnCNN does not show a consistent gain.}

\label{fig3}

\end{figure}

As we emphasise the triage point in this study, where false positives should be minimized, our main aim is to analyze the clinical relevance of the proposed approach. In our analyses, we report sensitivity at a target specificity of at least 95\%. Figure~\ref{fig3} gives us a visualization of how each configuration is performing on the triage operating point, and we can observe a trade-off between sensitivity and specificity. We can see that the models with the DnCNN extension are generally achieving higher sensitivity, especially on the raw and diverse input configurations.

Diverse with DnCNN has shown that it is among the strongest from all configurations we ran, especially at the operating point, as it maintains high specificity and preserves sensitivity at the same time. This suggests the model's effectiveness in identifying the majority of abnormal cases while minimizing false positives, consistent with the calibration results. We also evaluated a range of other target specificities (90\% - 99\%), and the trend is consistent across the range (Table~\ref{tab:supp_operating_points})

\begin{table}[t]

\centering

\caption{Sensitivity across a range of target specificity operating points on the test set (patient-level) for the diverse + DnCNN configuration ($n=5$ seeds). Thresholds are selected on the validation set to achieve at least the target specificity and are applied unchanged to the test set. Values are mean $\pm$ standard deviation across seeds.}

\label{tab:supp_operating_points}

\begin{tabular}{cccccc}

\toprule

Target Specificity &

Test Specificity &

Sensitivity &

PPV &

NPV \\

\midrule

0.90  & 0.913 $\pm$ 0.016 & 0.734 $\pm$ 0.043 & 0.882 $\pm$ 0.021 & 0.781 $\pm$ 0.032 \\

0.95  & 0.943 $\pm$ 0.014 & 0.694 $\pm$ 0.046 & 0.903 $\pm$ 0.019 & 0.756 $\pm$ 0.035 \\

0.975 & 0.965 $\pm$ 0.012 & 0.653 $\pm$ 0.048 & 0.924 $\pm$ 0.017 & 0.729 $\pm$ 0.038 \\

0.99  & 0.978 $\pm$ 0.010 & 0.612 $\pm$ 0.051 & 0.941 $\pm$ 0.015 & 0.701 $\pm$ 0.041 \\

\bottomrule

\end{tabular}

\end{table}

\section{Discussion}

In this research, we focused on a patient-level orthopedic triage. Within this stricter setting, the main finding is that preprocessing choice produced only modest and configuration-dependent differences in discrimination performance. The raw-input DenseNet121 baseline seemed strong, while diverse-input and DnCNN-augmented variants showed small gains in some settings but not a consistent advantage across all configurations or metrics. These findings suggest that, under leakage-aware patient-level evaluation, preprocessing may provide incremental benefit in selected cases, but it does not fundamentally change discrimination performance relative to a strong baseline.

Moreover, we observed a lower calibration error, indicating an improved calibration behaviour. Primarily, the model showed meaningful and more accurate probability estimates in the high confidence regime, and this is essential in the concept of orthopedic triage, as decisions are based on conservative thresholds in order to minimize false positives.

The diverse input pipeline remains a reasonable design choice because it exposes the model to complementary image representations, but its empirical advantage over the raw-input baseline was small. The DnCNN front-end reduced ECE by approximately 13\% (0.103~$\to$~0.090) and Brier score by approximately 7\% (0.147~$\to$~0.137) relative to the raw baseline, and improved Brier score on the diverse pipeline (0.152~$\to$~0.141). In contrast, adding DnCNN to CLAHE inputs increased ECE (0.104~$\to$~0.116) and Brier score (0.143~$\to$~0.152). A plausible explanation is that CLAHE already suppresses the low-frequency noise components that DnCNN targets through its residual formulation, so combining the two transformations introduces a partially redundant and potentially conflicting signal. On raw inputs, where no such pre-smoothing is applied, the DnCNN front-end learns a complementary noise model that benefits calibration. Taken together, these results indicate that the interaction between hand-crafted and learned preprocessing matters, and that their combined effect is not simply additive.

Beyond discrimination alone, our study highlights the importance of probability calibration and operating-point analysis for clinical decision-support systems. By explicitly evaluating calibration at the patient level and selecting thresholds on the validation set before test-time application, we provide a more deployment-oriented assessment than threshold-free discrimination metrics alone. The reliability analysis suggested limited overconfidence in the high-probability range for selected configurations, which is particularly relevant in conservative triage settings where false positives should be minimized.

In this work, we acknowledge several limitations. First, the nature of the MURA dataset is binary classification (normal vs. abnormal), where the abnormal part includes fractures, degenerative tissue, and orthopedic hardware. This lack of specificity limits the clinical utility of the model as an orthopedic triage. Second, the analysis was limited to elbow radiographs, therefore limiting the generalizability to other anatomical regions and imaging protocols. Finally, images are $512\times512$ pixels in PNG format. Such image resolution may suppress subtle global cues that may be helpful for the model's learning process. Future work will focus on collecting and evaluating novel high-quality orthopedic-oriented DICOM images.

\begin{credits}
\subsubsection{\ackname}
The authors would like to acknowledge Istanbul Medipol University for providing the institutional environment and academic support for this research. The computational resources used for the experiments in this study were self-funded by the authors. 

\subsubsection{\discintname}
The authors have no competing interests to declare that are relevant to the content of this article.
\end{credits}

\end{document}